\documentclass[sigconf]{acmart}
\AtBeginDocument{%
  }
\usepackage{listings}
\usepackage{enumitem}
\usepackage{threeparttable}
\usepackage{nopageno}
\setcopyright{acmlicensed}
\copyrightyear{2025}
\acmYear{2025}
\setcopyright{acmlicensed}
\acmConference[MRAC '25] {Proceedings of the 3rd International Workshop on Multimodal and Responsible Affective Computing}{October 27--31, 2025}{Dublin, Ireland.}
\acmBooktitle{Proceedings of the 3rd International Workshop on Multimodal and Responsible Affective Computing (MRAC '25), October 27--31, 2025, Dublin, Ireland}
\acmISBN{979-8-4007-2052-9/2025/10}
\acmDOI{10.1145/3746270.3760225}
\begin{document}
\title{ECMF: Enhanced Cross-Modal Fusion for Multimodal Emotion Recognition in MER-SEMI Challenge}
\author{Juewen Hu}
\email{hujuewen@bigai.ai}
\orcid{0009-0007-0083-5736} 
\affiliation{%
  \institution{State Key Laboratory of General Artificial Intelligence, BIGAI}
  \city{Beijing}
  \country{China}
}
\author{Yexin Li}
\email{liyexin@bigai.ai}
\orcid{0000-0001-5907-8978} 
\affiliation{%
  \institution{State Key Laboratory of General Artificial Intelligence, BIGAI}
  \city{Beijing}
  \country{China}
}
\author{Jiulin Li}
\email{lijiulin@bigai.ai}
\orcid{0009-0004-1578-5485}
\affiliation{%
  \institution{State Key Laboratory of General Artificial Intelligence, BIGAI}
  \city{Beijing}
  \country{China}
}
\author{Shuo Chen}
\email{chenshuo@bigai.ai}
\orcid{0000-0002-0257-1240} 
\affiliation{%
  \institution{State Key Laboratory of General Artificial Intelligence, BIGAI}
  \city{Beijing}
  \country{China}
}
\author{Pring Wong}
\email{huangping@bigai.ai}
\thanks{*Corresponding Author}
\authornotemark[1] 
\orcid{0000-0003-1752-599X} 
\affiliation{%
  \institution{State Key Laboratory of General Artificial Intelligence, BIGAI}
  \city{Beijing}
  \country{China}
}
\renewcommand{\shortauthors}{Juewen Hu, Yexin Li, JiuLin Li, Shuo Chen, Pring Wong}

\begin{abstract}
  
  Emotion recognition plays a vital role in enhancing human-computer interaction. In this study, we tackle the MER-SEMI challenge of the MER2025 competition by proposing a novel multimodal emotion recognition framework. To address the issue of data scarcity, we leverage large-scale pre-trained models to extract informative features from visual, audio, and textual modalities. Specifically, for the visual modality, we design a dual-branch visual encoder that captures both global frame-level features and localized facial representations. For the textual modality, we introduce a context-enriched method that employs large language models to enrich emotional cues within the input text. To effectively integrate these multimodal features, we propose a fusion strategy comprising two key components, i.e., \textit{self-attention mechanisms} for dynamic modality weighting, and \textit{residual connections} to preserve original representations. Beyond architectural design, we further refine noisy labels in the training set by a multi-source labeling strategy. Our approach achieves a substantial performance improvement over the official baseline on the MER2025-SEMI dataset, attaining a weighted F-score of 87.49 \% compared to 78.63 \%, thereby validating the effectiveness of the proposed framework.
  
\end{abstract}
\ccsdesc[500]{Human-centered computing~Human computer interaction (HCI)}
\ccsdesc[500]{Computing methodologies~Artificial intelligence~Natural language processing}
\ccsdesc[500]{Computing methodologies~Artificial intelligence~Computer vision}
\ccsdesc[500]{Computing methodologies~Machine learning~Multimodal learning}

\keywords{Multimodal emotion recognition, Transformer architecture, Self-attention mechanism, Large language models, Computer vision, Natural language processing, Audio processing, MER2025-SEMI dataset, Cross-modal fusion, Deep learning}
\maketitle

\section{Introduction}
Artificial intelligence (AI) has revolutionized numerous industries, with growing emphasis on enhancing its anthropomorphic capabilities. A fundamental aspect of this endeavor is equipping AI systems with the ability to understand human emotions, which is critical for effective human-computer interaction (HCI). Accurate emotion recognition can greatly improve user experience and elevate the quality of interaction \cite{Picard97}.

As a subtask of the MER2025 competition \cite{MER25}, the MER-SEMI challenge seeks to advance the field of emotion recognition by providing a semi-supervised learning setting that includes both labeled and unlabeled video data\cite{Lian2023}\cite{Lian24}. Its objective is to classify each video sample into one of six predefined emotion categories, i.e., \textit{worry}, \textit{happiness}, \textit{neutral}, \textit{anger}, \textit{surprise}, and \textit{sadness}.

However, emotion recognition poses several significant challenges. Recognizing emotions from video involves multiple modalities, including text, visual, and audio. The core difficulties lie not only in effectively encoding and extracting informative features from each modality but also in integrating these heterogeneous signals for accurate classification. Moreover, the scarcity of labeled data further complicates the task. For instance, MER2025 provides only 7,369 labeled samples\cite{Lian2025}, which limits the ability to train fully supervised models and increases the reliance on semi-supervised or pre-trained approaches \cite{Poria17}.

To address the issue of data scarcity, we leverage pre-trained models as feature extractors, which have demonstrated strong generalization capabilities in data-scarce scenarios. These models, trained on large-scale corpora, provide robust and transferable representations for each modality. For the textual modality, BERT \cite{Devlin19} and RoBERTa \cite{Liu19} capture rich semantic and syntactic information through contextualized embeddings. In the visual domain, I3D \cite{Carreira17} and SlowFast \cite{Feichtenhofer19} extract both spatial and temporal features to effectively represent dynamic expressions and motion cues. For the auditory modality, Wav2Vec \cite{Baevski20} and HuBERT \cite{Hsu21} produce expressive speech representations capable of capturing variations in tone, pitch, and prosody. Furthermore, to further enhance performance, we propose a dual-branch visual encoder and a context-enriched method for the visual and textual modalities, respectively, both built upon the corresponding pre-trained models.

In addition, we design a fusion strategy to effectively integrate the rich features extracted from multiple modalities. Prior studies have shown that conflicting or redundant signals across modalities can degrade performance, highlighting the importance of balancing each modality’s contribution in multimodal emotion recognition. To address this, rather than directly concatenating features, we employ attention mechanisms to dynamically weight the importance of each modality. This approach enhances the quality of the joint representation and promotes more robust and accurate emotion classification \cite{Vaswani17}.

\begin{figure*}[ht]
  \centering
  \includegraphics[width=\textwidth]{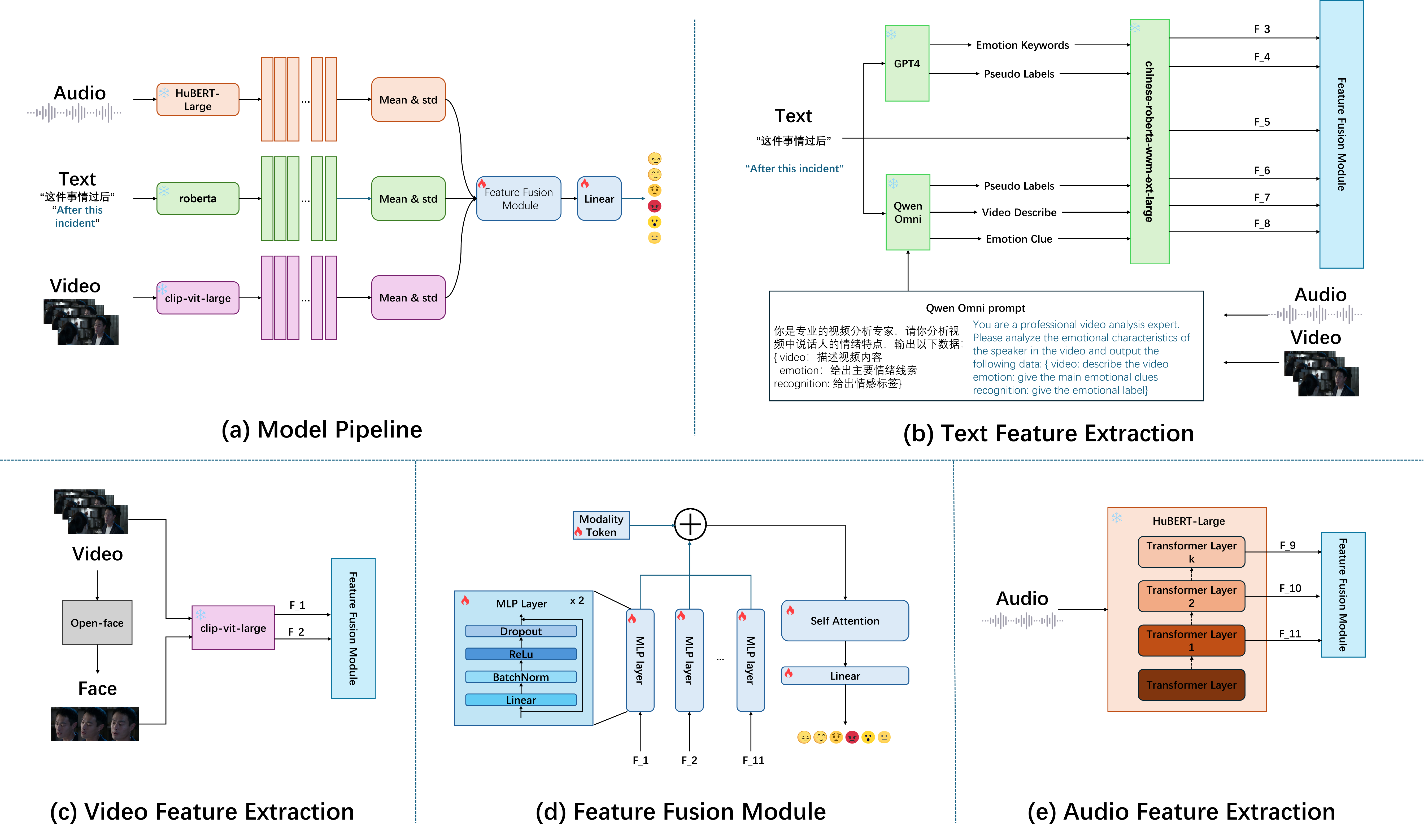}
  \caption{
  Enhanced Cross-Modal Fusion Architecture for Multimodal Emotion Recognition. 
  (a) \textbf{Model Pipeline}: integrated workflow comprising data input, feature extraction, feature fusion module, and classification. 
  (b) \textbf{Text Feature Extraction}: context-enriched encoding via Chinese-RoBERTa-wwm-ext-large, enhanced with GPT4-generated keywords and Qwen-omni emotion clues. 
  (c) \textbf{Video Feature Extraction}: dual-branch encoding with OpenFace for facial detection, and CLIP-ViT-Large for spatial encoding of full frames and facial regions. 
  (d) \textbf{Feature Fusion Module}: multimodal integration through residual connections, Modal\_Token incorporated, and two-layer self-attention.
  (e) \textbf{Audio Feature Extraction}: prosodic feature extraction using layers 16-21 of HuBERT-Large.
  \textit{Note: F\_x denotes feature streams from respective modalities.}
  }
  \Description{
  A diagram of an enhanced cross-modal fusion architecture for emotion recognition, showing data preprocessing, feature extraction (text: 6 streams, video: 2 streams, audio: 3 streams), fusion, and classification.
  }
  \label{fig:arch}
\end{figure*}

Beyond the proposed model architecture, we further refine noisy labels in the training set through a multi-source labeling strategy. Specifically, we train weak classifiers on each individual modality using the original training data. For each sample, we then collect emotion labels from the weak classifiers as well as from a large language model (LLM). The final refined label is determined via majority voting across these sources. To ensure label reliability, a small subset of samples exhibiting highly inconsistent predictions is manually reviewed and corrected as necessary. 

In summary, this study proposes a multimodal emotion recognition framework to address the MER2025 challenge. Our contributions are threefold.

\begin{itemize}[leftmargin=2em]
\item To address the issue of data scarcity, we leverage appropriate pre-trained models as multimodal feature extractors. Specifically, for visual modality, we design a dual-branch visual encoder that captures both global frame-level features and localized facial representations. For textual modality, we propose a context-enriched method using LLMs to enrich emotional cues in the text inputs.

\item To handle modality competition, we design a fusion strategy that dynamically weights different modalities to ensure robust performance.

\item Extensive experiments conducted on the official dataset demonstrate a significant improvement over the baseline, achieving a weighted F-score of 87.49 \% compared to 78.63 \%.
\end{itemize}

\section{Related Works}
\paragraph{\textbf{Modality competition in multimodal fusion}} Studies have shown that different modalities, such as audio, video, and text, may compete during fusion, adversely affecting emotion recognition performance. Huang et al. \cite{Huang22} investigated the reasons for failures in joint training of multimodal networks, emphasizing the importance of balancing contributions from each modality. Katak et al. \cite{Khattak23} proposed the Maple method, which adaptively focuses on relevant modalities through prompt learning to mitigate competition. Lian et al. \cite{MER25} addressed noise and open-vocabulary scenarios in semi-supervised learning, making their approach suitable for real-world applications.

\paragraph{\textbf{Spatiotemporal features of video}} Video-based emotion recognition requires capturing both spatial and temporal dynamics. Ruan et al. \cite{Wang17} utilized 3D CNNs to extract spatiotemporal features from audio and video, improving recognition accuracy. Another study \cite{Cho20} applied 3D CNNs to model spatiotemporal representations in EEG signals, achieving significant results. Deep learning methods, such as 3D CNNs and LSTMs, excel at learning complex patterns, making them well-suited for dynamic emotion analysis.

\section{Methodology}
This section elaborates on the proposed method, which is organized into three subsections. First, we present the overall framework and provide a high-level description of its architecture. Second, we detail the feature extraction process for each modality. Finally, we describe the multimodal fusion strategy employed to integrate the extracted features.

\subsection{Model Architecture}
We propose a multimodal emotion recognition framework, whose overall architecture is illustrated in Figure~\ref{fig:arch} (a). The framework consists of three main components, i.e., data input, feature extraction, and feature fusion.

At the data input stage, we first extract each modality from the raw video data. Modality-specific features are then obtained using three pre-trained models, i.e., HuBERT-Large for audio, Chinese-RoBERTa-wwm-ext-large for text, and CLIP-ViT-Large for visual information. The extracted feature vectors are subsequently standardized. Finally, each modality’s representation is fed into a dedicated Feature Fusion Module to generate the final multimodal representation for subsequent emotion classification.

\subsection{Feature Extraction}

\subsubsection{Audio}
Speech plays a crucial role in emotion recognition, as identical content conveyed with different intonations can express distinct emotions. Therefore, extracting audio features such as pitch, volume, and tone is essential. Various pre-trained encoders differ in their capability to capture such features.

Inspired by prior work \cite{Zhao24}, we employ HuBERT-Large to extract emotional features from audio signals. Specifically, we utilize the outputs from layers 16 to 21 of the HuBERT-Large model, as these layers have been shown to capture richer prosodic and spectral patterns \cite{Zhao24}. Their representations exhibit enhanced adaptability to acoustic variations, making them particularly effective for emotion recognition. As illustrated in Figure~\ref{fig:arch} (e), the audio data is fed into the HuBERT-Large model, from which the selected layer outputs are standardized and passed to the feature fusion module.

\subsubsection{Text}
Textual content plays a pivotal role in conveying emotional expressions, as it captures both the contextual background and causal relationships of events depicted in videos. Emotion-related lexical elements within the text, such as sentiment-bearing words or specific emotional particles, greatly contribute to distinguishing between different emotional states. However, the emotion recognition accuracy of the text modality often lags behind that of the audio and visual modalities.

To address this limitation, we propose a context-enriched method that leverages LLMs to enhance emotional cues within textual inputs. Specifically, we augment the original text using GPT-4 and Qwen-Omni \cite{Qi24}. GPT-4 is employed to generate pseudo-labels and emotion-related keywords for each text sample, thereby enriching the emotional context \cite{OpenAI23}. In parallel, Qwen-Omni processes audio and visual content using carefully designed prompts, as shown in Figure~\ref{fig:arch} (b), generating pseudo-labels, detailed video descriptions, and auxiliary emotional cues \cite{Xu25}. These enriched outputs, along with the original text inputs, are subsequently encoded by the Chinese-RoBERTa-wwm-ext-large model to produce enhanced textual features. The resulting embeddings are standardized and integrated into the feature fusion layer for downstream analysis.

\subsubsection{Video}
Human expressions and body language serve as key indicators of emotion. The MER2025 baseline \cite{MER25} extracts and encodes facial regions from each video frame, achieving moderate performance. Recognizing that body movements also contribute significantly to emotional expression, we develop a dual-branch visual encoder that integrates both global frame-level features and facial representations.

As illustrated in Figure~\ref{fig:arch} (c), for each video frame, facial information is detected and extracted using OpenFace \cite{Baltrusaitis16}. Both the extracted facial patches and the full video frames are then fed into the CLIP-ViT-Large encoder \cite{Radford21}. The resulting dual-scale visual features are standardized and subsequently fed into the feature fusion layer.

\subsection{Feature Fusion}
Multimodal feature fusion plays a pivotal role in emotion recognition, as it enables the effective integration of emotional cues from different modalities. Although modality-specific features are extracted using appropriate pre-trained models—HuBERT-Large for audio, Chinese-RoBERTa-wwm-ext-large for text, and CLIP-ViT-Large for video—some emotion-irrelevant information may still be retained in the feature representations. Therefore, a robust fusion strategy is essential to refine these representations and emphasize emotionally salient information.

We propose a fusion method based on self-attention mechanisms with residual connections. As illustrated in Figure~\ref{fig:arch} (d), the outputs from each feature extractor are first processed through an encoder incorporating a residual module, which preserves original information while capturing additional emotional cues and projecting features into a unified space. Standardization and dropout layers are applied to accelerate model convergence.

For each modality, a learnable \lstinline{Modal_Token} is prepended to the feature sequence to encode modality-specific information—such as distinguishing between audio, text, or video features—analogous to the use of positional encodings in Transformers. The resulting sequence is then fed into two self-attention layers, which produce the final emotion prediction \cite{Vaswani17}.

\subsection{Implementation Details}
Beyond the architectural design, we further enhance practical performance by refining noisy labels in the training set. In addition, we employ ensemble learning to determine the final emotion label for each sample, thereby improving label reliability.

\subsubsection{Refining Noisy Labels}
During data preprocessing, we observed inconsistencies between certain training labels and their corresponding video content in the MER2025-SEMI dataset. To address this issue, we refine the noisy labels using a multi-source labeling strategy. Specifically, we trained weak classifiers on each individual modality using the original training data. For each sample, we collected emotion label predictions from these weak classifiers. Additionally, we leveraged Qwen-Omni to generate auxiliary emotion labels. We then applied a majority voting scheme across all sources to derive refined labels. For samples where all predicted labels disagreed with the original annotation, we manually verified and corrected the labels to ensure quality. This relabeling improved model performance, consistent with findings in prior work \cite{Malik24}.

\subsubsection{Ensemble Learning}
Based on the architecture illustrated in \autoref{fig:arch}~(a), we construct several model variants to enhance label quality through ensemble learning. Specifically, we either randomly remove certain modules from the original framework or train the same model using different random seeds to introduce diversity. These variant models are then used to predict emotion labels for each video sample. Finally, we apply a majority voting scheme across the predictions from all variants to obtain the final ensemble-based emotion labels.

\section{Experiments}
This section presents the experimental setup and results of the proposed framework, including details on the dataset, hyperparameter configurations, and performance evaluation.

\subsection{Dataset}

We utilized the MER2025-SEMI dataset, comprising 7,369 labeled samples and 20,000 unlabeled samples. The official baseline employs five-fold cross-validation to split the training set into training and validation subsets, averaging the best results across the five validation sets to obtain the final weighted F-score (WAF).

\subsection{Settings}

To ensure stable training, we set the hidden dimension to 128, the dropout rate to 0.6, and use two self-attention heads, with gradient clipping at 1.0, a learning rate of 5e-5, and up to 200 training epochs.


In addition to comparing our method with the official baseline, we conduct studies to evaluate the contribution of each module in our framework. While several components have been clearly explained in previous sections, we provide further clarification for the less intuitive ones as follows.

\begin{itemize}[leftmargin=2em]
\item \textbf{Norm} standardizes features using the mean and standard deviation to ensure consistent distributions across modalities.
\item \textbf{Fold-6} applies 6-fold cross-validation to improve generalization by training and validating on six different data splits.
\item \textbf{GPT4-label} leverages GPT-4 to analyze video content and generate emotion labels, thereby enhancing the quality of text-based feature representations.
\item \textbf{GPT4-keywords} utilizes GPT-4 to extract semantic keywords from textual data, enriching the text inputs.
\item \textbf{MLP} refines the multilayer perceptron architecture to re-encode features into a unified space.
\end{itemize}

\begin{table}
\caption{WAF of Different Methods.}
\label{tab:f1_scores}
\begin{threeparttable}
\begin{tabular}{lcc}
\toprule
Methods & WAF / val & WAF / test \\
\midrule
Baseline & 82.05\% & 76.80\% \\
+ Multi-source labeling strategy & 82.31\% & 78.67\% \\
+ Dual-branch visual encoder & 82.80\% & 78.68\% \\
+ Modal\_Token & 83.27\% & 78.84\% \\
+ Norm & 83.20\% & 84.40\% \\
+ Roberta & 83.50\% & 85.30\% \\
+ Fold-6 & 83.60\% & 85.60\% \\
+ GPT4-label & 84.09\% & 86.08\% \\
+ GPT4-keywords & 84.15\% & 86.49\% \\
+ MLP & 84.29\% & 86.94\% \\
+ Selective Hubert\_Layer & 84.84\% & 87.14\% \\
+ Ensemble learning & - & 87.49\% \\
\bottomrule
\end{tabular}
\end{threeparttable}
\end{table}

\subsection{Results}
Our results, as shown in Table~\ref{tab:f1_scores}, demonstrate that the proposed method substantially outperforms the baseline. Although the baseline achieves comparable performance on the validation set, its test performance drops to 76.8 \%, which is even lower than the 78.63 \% reported in the official benchmark paper\cite{Lian2025}.

At the data level, our multi-source labeling strategy leads to more accurate labels and improved model generalization compared to the baseline. At the feature level, the dual-branch visual encoder enables complementary integration of global scene information and fine-grained facial cues, effectively boosting visual representation quality. For the text modality, the incorporation of LLMs enriches emotional context, thereby mitigating its relative underperformance. In the audio modality, selectively utilizing emotion-sensitive layers from HuBERT-Large further strengthens emotional feature extraction. Finally, ensemble learning consistently boosts performance, yielding gains of 0.5–1.3 percentage points.

\section{Conclusion}

This study presents a multimodal emotion recognition framework for the MER2025-SEMI challenge, leveraging pre-trained models and advanced fusion techniques to enhance performance under limited labeled data. Our contributions include: a context-enriched method using LLMs to improve the emotional expressiveness of text features; a dual-branch visual encoder integrating global frame-level features and localized facial representations to enhance visual modality analysis; a fusion strategy based on self-attention with residual connections to effectively integrate multimodal features; and a multi-source labeling strategy to correct noisy labels in the training set. Experimental results demonstrate superior performance on the MER2025-SEMI dataset, significantly outperforming the baseline. Future work will explore additional data augmentation and fusion strategies to further enhance the accuracy and robustness of the proposed emotion recognition framework.

\begin{acks}
  This work was supported by the State Key Laboratory of General Artificial Intelligence, BIGAI. We thank our colleagues for their valuable feedback during the development of this project.
\end{acks}

\bibliographystyle{ACM-Reference-Format}
\balance
\bibliography{sample-base}


\end{document}